\documentclass{article}

\PassOptionsToPackage{numbers, compress}{natbib}
\usepackage[final]{midl_2018}

\usepackage[utf8]{inputenc} 
\usepackage[T1]{fontenc}    
\usepackage{hyperref}       
\usepackage{url}            
\usepackage{booktabs}       
\usepackage{amsfonts}       
\usepackage{nicefrac}       
\usepackage{microtype}      

\usepackage[noblocks]{authblk}   

\usepackage{booktabs}       
\usepackage{graphics}        

\usepackage{graphicx}	
\usepackage[]{algorithm2e}
\usepackage[table,x11names]{xcolor} 

\usepackage[table,x11names]{xcolor} 
\definecolor{lightgray}{rgb}{0.95,0.95,0.95}

\title{Training of a Skull-Stripping Neural Network with efficient data augmentation}

\usepackage{authblk}
\begin{document}

\author[1]{\textbf{Gabriele Valvano} \thanks{Corresponding author: \emph{gabriele.valvano@imtlucca.it} (G. Valvano).} }
\author[2]{\textbf{Nicola Martini}}
\author[1]{\textbf{Andrea Leo}}
\author[2,3]{\textbf{Gianmarco Santini}}
\author[2]{\textbf{Daniele Della Latta}}
\author[1]{\textbf{Emiliano Ricciardi}}
\author[2]{\textbf{Dante Chiappino}}

\affil[1]{IMT School for Advanced Studies Lucca, Lucca, Italy}
\affil[2]{Imaging department, Fondazione Gabriele Monasterio, Massa, Italy}
\affil[3]{Department of Information Engineering, University of Pisa, Pisa, Italy}
\renewcommand\Authands{ and }  



\maketitle

\begin{abstract}
Skull-stripping methods aim to remove the non-brain tissue from acquisition of brain scans in magnetic resonance (MR) imaging. Although several methods sharing this common purpose have been presented in literature, they all suffer from the great variability of the MR images. In this work we propose a novel approach based on Convolutional Neural Networks to automatically perform the brain extraction obtaining cutting-edge performance in the NFBS public database. Additionally, we focus on the efficient training of the neural network designing an effective data augmentation pipeline. Obtained results are evaluated through Dice metric, obtaining a value of 96.5\%, and processing time, with 4.5s per volume. 
\end{abstract}

\section{Introduction}

Skull-stripping is the automatic extraction of brain tissue from magnetic resonance imaging (MRI) and is usually the first step in most neuroimaging pipelines. For this reason its results affect many operations subsequently performed on anatomical MR images, including segmentation and registration to standard space. In this context, the development of accurate and robust methods for the identification of brain tissue  is a fundamental, yet challenging task, due to the great variability in the brain morphology across subjects and to the variability in voxel intensity obtained through different MR scanners, coils and acquisition protocols.

Several skull-stripping methods have been proposed and implemented in many different software packages. Although many of these methods achieve high performance levels on most MR structural images, they may fail to produce satisfactory results when the acquisition conditions or study populations change.

To address some of the problems faced by classical algorithms in the skull-stripping task, in this paper we propose the use of a non-linear approach based on Convolutional Neural Networks (CNNs). 

CNNs are a class of deep learning algorithms that, in the last few years, have become the state of the art for semantic segmentation and many other computer vision tasks. Their strength derives from the capability to avoid the direct definition of the image characteristics to analyze, automatically defining these salient features by iteratively minimizing a cost function. 

Furthermore, although the training phase of these algorithms is usually slow, the testing process is rather rapid and allows segmentation masks to be obtained much more quickly with respect to the most common approaches.

However, the efficient training of CNNs often demands a considerable amount of high quality and wide variety of data to gain a good generalization. The underlying idea is: the more data we have, the better the learning algorithm will work; but at the same time we can not sacrifice the good quality of the data, as training CNNs with low-quality data will lead in a worse accuracy of the algorithm, which will be working more coarsely.

Unfortunately every data collection process is associated with a cost, which can be in terms of time, money, human effort and computational resources. For this reason public database are usually relatively small and do not contain a sufficient variability of data. In fact, MR brain scans are usually performed on a reduced number of different MR scanners and with only a few of the possible acquisition protocols. Therefore, training a machine learning algorithm to perform brain extraction on these dataset could bring to a low generalization of the model on different images.

Having a small dataset of high quality samples, a common workaround for the lack of data is the generation of new artificial samples starting from the available data resources: this process is referred to as data augmentation and aims to compensate for the cost involved in further data collection and labeling. There are several ways to augment the data, such as matrix rotation, mirroring and translation.

In this work we trained a convolutional neural network with a relatively small number of brain volumes (90 in total), preferring the usage of a number of high quality hand-made brain masks with respect to suboptimal automatic segmentations. In particular we trained the neural net to automatically extract the brain tissue from the MRI images, focusing on how to avoid the network from overfitting even in the case of a reduced number of available data, achieving a good generalization of the model. 

During the study we employed a CNN architecture which is a modified version of the well-known U-Net \cite{ronneberger2015u}, which is one of the most popular architectures used for semantic segmentation. We also analyzed the contribute of several types of data augmentation in the final performance of the algorithm.

\section{State of the art}

The importance of having an effective brain extraction tool in neuroimaging is highlighted by the quantity of papers suggesting different strategies to perform this segmentation. Since T1-weighted MRI images have a good contrast, they are the most common structural MRI acquired in neuroimaging studies. Consequently, most of the brain extraction methods are designed to work with T1-weighted images. The most popular skull-stripping methods have publicly available implementations and are designed to work mainly on T1-weighted data. 

A short list of the most used algorithms follows:

\begin{itemize}

\item Brain Extraction Tool (BET) \cite{smith2002fast}. Part of the FSL package, it is widely used. It employs a deformable model which, starting from a sphere placed around the center of gravity of the brain, subject to a force field, expands to adapt to the surface of the brain. BET is very fast and relatively insensitive to parameter settings. It also provides good results, although it often produces false positives. This can be solved if, after running BET one time, we use the preliminary mask obtained to register the brain to an atlas. Then the atlas mask is used to drive a second BET application. The main disadvantage of this process is that the registration turns the method much slower.

\item Brain Surface Extraction (BSE) \cite{shattuck2001magnetic}. Even the BSE method is divided into a series of consequential steps: the filtering of anisotropic diffusion, the detection of margins and a chain of morphological operations. Although BSE can provide high quality brain segmentations, it often requires careful parameters tuning for specific images.

\item 3dSkullStrip. It is part of the AFNI package \cite{ref_afni} and consists of a modified version of BET. Similarly to BET, it uses the spherical surface expansion approach, but It also includes some precautions to avoid eyes and ventricles, while preserving the brain tissue with appropriate adjustments.

\item A very popular public method is the hybrid approach of \cite{segonne2004hybrid}, available in the FreeSurfer software package \cite{ref_FreeSurfer}. It combines watershed algorithms and deformable surface models. The watershed algorithm builds an initial estimate of the brain volume based on the three-dimensional connectivity of the white matter. A statistical atlas, generated from a set of accurately segmented brains, is used to refine the mask in order to correct the segmentation. The latest version of FreeSurfer employs GCUT to refine the output.

\item ANTS \cite{avants2011reproducible}. The first step of the skull-stripping pipeline using ANTs involves an iterative process consisting in the construction of an optimal template starting from structural MRI data. This construction iterates between the current estimation of the template and the registration of each subject on it. A probabilistic segmentation mask of the brain is subsequently generated by deforming an existing mask to match the obtained template. Then, after further morphological refinements, Atropos is used to generate an initial estimate of the segmentation of the three tissues inside the brain mask. Each of these three sub-masks incurs into a series of separate morphological operations and is later combined with the others in order to create the final brain mask. This mask is finally refined by new operation of dilation, erosion and filling.
  
\item ROBEX \cite{iglesias2011robust}. The method employs a discriminative and a generative model to extract the brain tissue. The discriminative model is a Random Forest classifier trained to detect the boundary regions of the brain, while the generative model is meant to ensure the plausibility of the result. Afterwards, a refining of the contour is pursued using graph cuts to obtain the final segmentation mask.

\end{itemize}

\section{Material and Methods}
\subsection{Data}\label{subsect_data}

The Neurofeedback Skull-stripped (NFBS) repository \cite{ref_nfbs} is an online database of T1-weighted anatomical MRI scans. The repository contains data from 125 participants, aged 21 to 45, with a variety of clinical and subclinical psychiatric symptoms. The images, acquired on a Siemens Trio 3T scanner, have a spatial resolution of 1 $mm^3$ and a matrix size of 256x256x192 voxels. A manually-corrected segmentation mask is released alongside each individual image.

In order to determine which volumes should be assigned to train, validation and test sets, we first performed the automatic skull-stripping with all the above mentioned classical approaches; then we built a performance ranking of the subject volumes when segmented with a certain algorithm; for each scan, we evaluated the error of each method with the same metric. Therefore, to detect the most difficult volumes to segment inside the dataset, we added up the partial rankings obtained from each brain extraction method, so to obtain the overall score for each subject. 

We used as test samples the 30 volumes with the lowest segmentation quality, the following worse 5 to construct the validation set and the remaining 90 as training volumes. In this way, we could evaluate the algorithm performance in the case of very hard to segment input after training on easier cases.

As preprocessing, each volume was standardized before any further computation.

\subsection{Data augmentation}

From a preliminary analysis we experimented how, to some extent, it was possible for a network trained on bidimensional sagittal slices to correctly extract the brain tissue also on axial and coronal planes. This is most probably due to the ability of these algorithms to extract local features in order to have a more robust representation of the input, even in the case of relevant environment changes. For this reason, since when decontextualized and visualized on small windows the brain gyri look roughly the same in every projection, we decided to train the neural net on slices from the three projections, feeding in this way the model with very different images.

Starting from this observation we opted for training the CNN also on additional volume projections rather than only with the canonical coronal, sagittal and axial ones: this was possible through a random rotation carried out on the three-dimensional space of the volume and the subsequent slicing process on the usual axes.  

At this point a random roto-translation on plane for each slice was carried out separately. We subsequently applied a random shearing operation, in order to distort the image and alter the gyri structure. We highlight how, despite this process sometimes produces non plausible anatomy of the skull, it still helps increasing the variability of the brain tissue. Ultimately, by doing this, we wanted to encourage the network to recognize the brain and classify everything else as non-brain, regardless of its geometry and size.

We also augmented the data using flipping operations (i.e. up-down and left-right flipping) on the single slices and emulated a quadratic bias field by multiplying twice the input image for a randomly oriented linear mesh.

Finally, a small amount of random noise sampled from a uniform distribution was added to the image, as additional regularizer.

The amount of these operations is illustrated in Table \ref{table_data_aug_values}.


\begin{table*}[]
\centering
\resizebox{\columnwidth}{!}{

       \begin{tabular}{lcccccccc}  
           \cmidrule{1-7}
                & \begin{tabular}{@{}c@{}}Rotation 3D\\(x, y, z) (degree)\end{tabular} 
                & \begin{tabular}{@{}c@{}}Rotation\\(degree)\end{tabular}  	
                & \begin{tabular}{@{}c@{}}Translation\\(pixels)\end{tabular}  	
                & \begin{tabular}{@{}c@{}}Shearing factor\\(pixels)\end{tabular}  	
                & \begin{tabular}{@{}c@{}}Magnetic field\\ linear bias\end{tabular}   
                & \begin{tabular}{@{}c@{}}Random noise\end{tabular}   \\
           \midrule 
           \begin{tabular}{@{}c@{}}Amount\\range\end{tabular}   
           		&$(\pm 5, \pm5, \pm15)$  	
           		&$0 \div 360$  				
           		&$\pm20$  				
           		&$\pm0.10$ 					
           		&$0.5 \div 1.5$  	
           		&$0 \div 0.02$			\\
           \bottomrule
         \end{tabular} 
}
\caption{Data augmentation settings. If not specified, the operation is performed on a bidimensional slice.}
\label{table_data_aug_values}
\end{table*}

To evaluate the impact of the data augmentation operations on the final result, we tested the generalization capability of the learning algorithm when trained with a limited number of samples. In Table \ref{table_data_aug_settings} you can see the combination of operations we tested separately to see if they were able to bring benefits during the training process. To identify each combination, a unique label has been assigned. In particular, as can be seen from Label 2, we grouped the most common data augmentation operations under the same name, while we analyzed in more depth the other augmentation approaches.

In this circumstance the CNN was trained on a small minority of subjects to emphasize the contribution added by each operation. In particular, the training set consisted in the 20 volume scans which proved to be the easiest to segment with the classic approaches. The network training was pursued until convergence.

By training the CNN separately with each of the aforementioned operations we could evaluate each contribute.

\begin{table*}[]
\centering
\resizebox{\columnwidth}{!}{

       \begin{tabular}{lcccccccc}  
           \cmidrule{1-9}
              Label  & Rotation 3D 	  & Rotation 	& Translation 		& Shearing 	
              & \begin{tabular}{@{}c@{}}Left-right\\flipping\end{tabular}  	& \begin{tabular}{@{}c@{}}Up-down\\flipping\end{tabular}  	& \begin{tabular}{@{}c@{}}Magnetic\\field bias\end{tabular}   & \begin{tabular}{@{}c@{}}Random\\noise\end{tabular}   \\
           \midrule
         \textbf{0}    &$-$  					&$-$  				&$-$  								&$-$  				&$-$  				&$-$  	&$-$  	&$-$			\\
           		1    &$\checkmark$  	&$-$  				&$-$  								&$-$  				&$-$  				&$-$  	&$-$  	&$-$			\\
           		2    &$-$  					&$\checkmark$  &$\checkmark$  				&$-$  				&$\checkmark$  &$\checkmark$  	&$-$  	&$-$	\\
           		3    &$-$  					&$-$  				&$-$  				 	&$\checkmark$    				&$-$  				&$-$  	&$-$  	&$-$			\\
           		4    &$-$  					&$-$  				&$-$  				&$-$  	&$-$  	&$-$  				&$\checkmark$  	&$-$			\\
           		\textbf{ALL}    &$\checkmark$  	&$\checkmark$  	&$\checkmark$ 		&$\checkmark$  	&$\checkmark$  	&$\checkmark$  	&$\checkmark$ 	&$\checkmark$ 		\\
           \bottomrule
         \end{tabular} 
}
\caption{Tested data augmentation combinations. If not specified, the operation is performed on a bidimensional slice.}
\label{table_data_aug_settings}
\end{table*}

\subsection{Network architecture}

\begin{figure*}[t]
      \centering
          \includegraphics[width=1\textwidth]{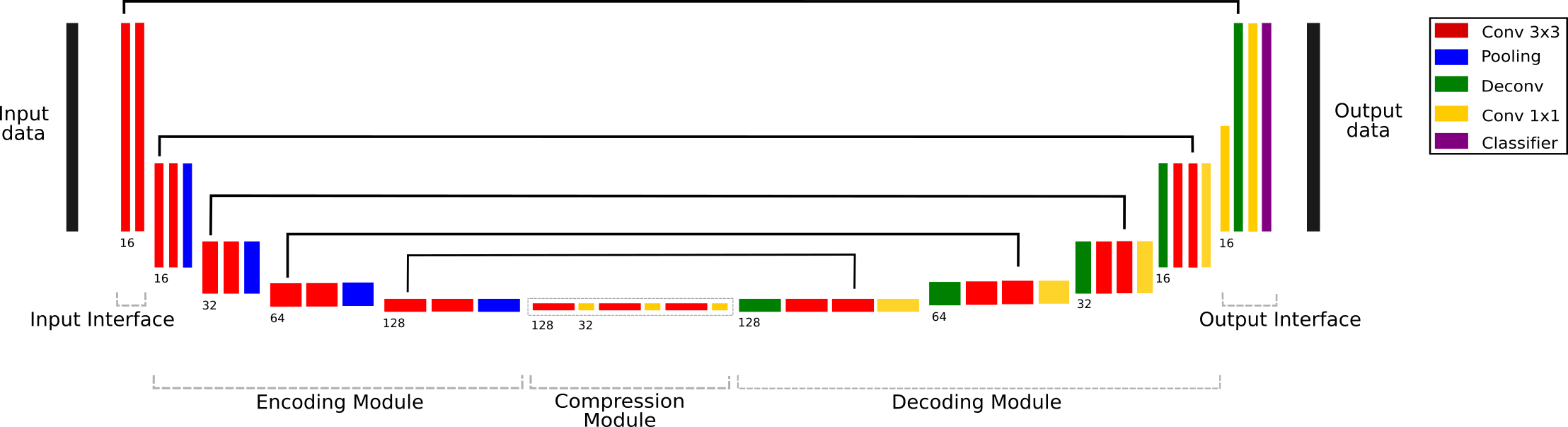}
      \caption{Neural network architecture. Notice that batch normalization layer are not rapresented in the image, but are still present after each convolutional and deconvolutional layer. Best viewed in color.}
      \label{fig_cnn_architecture}
\end{figure*}

IIn this work, we employed a modified version of the classical U-Net architecture \cite{ronneberger2015u} for the segmentation of the brain tissue (Figure \ref{fig_cnn_architecture}). 

Unless otherwise stated, each of the following mentioned convolutional layers employes a convolution of type 'same' with 3x3 kernels and ReLU activation function. Every layer is also followed by a batch normalization layer (not shown in the figure), which allows a consistent acceleration during the learning process, as suggested in \cite{ioffe2015batch}, by reducing the internal covariate shift. 

The deconvolutional layers use instead filter size and stride equal to 2.

Furthermore, every weight in the network is subject to the He initialization \cite{He2015delving} to lessen its dependence on the initial state. 

The number of channels for the convolutional layers is shown in the figure.


\subparagraph{The three main modules.}

As can be seen in Figure \ref{fig_cnn_architecture} the model consists of three main modules: the encoding module (EM), the compression module (CM) and the decoding module (DM). 

Since the EM corresponds to the contracting path of the network, it follows the typical architecture of a convolutional network. In detail, It consists of 4 repetitions of pairs of convolutional layers followed by a 2x2 max pooling operation. The usage of pooling layers here was particularly meant to decrease the computational burden of the network, while allowing a more abstract representation of the input, free from the constraint of small changes in the image. 

As the EM performs a progressive reduction of the image size, which is holded still throughout the compression module, in order to reconstruct the final segmentation mask we gradually enlarged the encoded matrices of features until we were able to match the original dimensions of the input. For this reason the decoding module was symmetrically built using repetitions of transposed convolutions for the up-sampling process and pairs of convolutional layers after each enlargement. Furthermore, we used "skip connections" to concatenate features maps from the encoding layers with the lower resolution ones of the decoding layers, obtaining a good localization and use of context while making easier the gradient flow during the backpropagation. Lastly, for each of these repetitions, we used an additional 1x1 convolutional layer to compress the number of channels of the features maps to the same number we would have had without the concatenation, so to compress the information and leave the following layer weights as numerous as in the encoding module. Note that with this filtering operation it is possible to build residual connections \cite{he2016deep}, if needed. In fact, 1x1 convolutions in theory allow the learning of the sum operation over the channels if this is the best option.

Another difference in our network compared to the canonical U-Net architecture is the addition of three pairs of 3x3 and 1x1 convolutional layers in the medial region of the network, constituting the compression module. The introduction of these layers aimed to perform a multi-step dimensionality reduction of the features maps preserving only the useful information to effectively reconstruct the final segmentation mask, while discarding the fine tissue-specific property. The purpose of this module was to cyclically expand and then compress the features maps building an analogous of the multilayer perceptron \cite{lin2013network}. Here, the 1x1 kernels bring a substantial reduction of the trainable parameters in the model, making it lighter and less prone to overfit, while allowing to learn a more non-linear function.

\subparagraph{Input and output interface.}

The input layers of the network basically correspond to a subsampling block. Here we have two 3x3 convolutional layers with stride 1 and then 2. In this way the output of this block results in a convolutional subsampling of the input.

As for the output layers, in order to respect the symmetry of the model we introduced an upsampling process. In particular, the output layers consist of: a convolutional layer employing 1x1 kernels to produce a 16-channel subsampled version of the output, then a deconvolutional layer to upscale it, a skip connection and the subsequent compression with the features maps produced by the first convolutional layer, the final classifier.

\subparagraph{Classification and loss function.}

For the classification process we implemented the softmax function. Here, for each pixel $y_i$ of the vectorized output matrix $Y$ the posterior probability to belong to the $l$-th class is computed by the softmax classifier, taking into account the vectorized input image $X$ and the network weight matrix $W$, as below:
\begin{eqnarray*}
	p( y_i=l | X) = \frac {e^{s_l}} {\sum_{k=1}^K e^{s_k}}    \qquad     l = 1, ..., K
\end{eqnarray*}
where $s$ is the score function which we define as $s = f(X, W)$, stated $f(\cdot)$ the non-linear function modelled by the neural net. With this formalism $s$ corresponds to the unnormalized log probabilities of the classes. 
			
For the training process we employed the Stochastic Gradient Descent algorithm in conjunction with Adam optimizer \cite{kingma2014adam} to minimize the Categorical Crossentropy:
\begin{eqnarray*}
	H( y, \bar{y} ) = -\sum_l  \bar{y_l} \cdot  log({y_l})
\end{eqnarray*}
where $\bar{y}_l$ is the ground truth label for the \textit{l}-th class and $y_l$ is the network output over the same class.
					
We addressed the problem of pixels class imbalance using a weighted loss function. The weights were selected according to the relative occurrences of each class in the training set, as:
\begin{eqnarray*}
	W_{loss_l} = 1 - {n_l \over N}		\qquad     l \in \{0, 1\}
\end{eqnarray*}
where $l$ corresponds to the one of the binary class of the segmentation, $n_l$ is the number of pixels with labels equal to $l$ and $N$ is the total number of pixels. 

We trained the network with batches of 16 slices, using a learning rate of $5*10^{-4}$.

\subparagraph{Test and brain extraction.}
Since the neural network has been trained on slices belonging to all three different projections, we can expect it to be able to extract a segmentation mask by working slice after slice along each axis. Therefore, we decided to evaluate the final mask as a weighted sum of the masks obtained along each projection in order to exploit the advantages that everyone brings. For example, since in the most apical slices the axial projection will be less reliable, these voxels will be much better classifiable based on the sagittal and coronal planes. The same concept applies to the outermost regions along the other axes.

Thus, the final mask is computed as:
\begin{eqnarray*}
mask_{final} = w_1*mask_1 + w_2*mask_2 + w_3*mask_3
\end{eqnarray*}
where the weights $w_1$, $w_2$, $w_3$ are evaluated as:
\begin{eqnarray*}
w_i = {N_i}/N_{tot} 		\qquad i \in \{1, 2, 3\}
\end{eqnarray*}
where $N_i$ is the number of ones in the slice on which the voxel resides along the $i$-th projection mask and $N_{tot}$ the total number of ones in all the projection masks.

These weights roughly reflect how many voxels of the brain are visible in the current slice for every projection. The less this number, the more peripherally we are and the less the current projection should be taken in consideration during the majority vote on the voxels classification.

Finally, an operation of labeling is performed to exclude possible false positives outside the brain mask and false negatives inside, while smoothing the boundary regions.

\subsection{Metrics}
To evaluate the performance of the model we employed the Dice overlap, defined as:
\begin{eqnarray*}
Dice = {2 |A \cap B| \over |A| + |B|}
\end{eqnarray*}

Dice was also used as reference index to evaluate the ranking of the volume scans when segmented by each classical approach (see Section \ref{subsect_data}).

We also evaluated the false positives rate (FPR) and false negatives rate (FNR) to have a clue of how many misclassified pixels we statistically have in the generated mask. These are defined as:

\begin{eqnarray*}
FPR = {\sum{False Positive} \over \sum{Condition Negative}}    \quad\quad     FNR = {\sum{False Negative} \over \sum{Condition Positive}}
\end{eqnarray*}

Additionally, we compared the results obtained on the test-set by using the most common skull-stripping methods with the one we suggest. 

A comparison in terms of average time spent for extracting each volume is also carried out. 

\section{Results}
Table \ref{table_data_aug_comparison} shows the contribute of each data augmentation process to the generalization of the neural network when trained on a reduced training set. In Table \ref{table_single_vs_all_projs} can be seen the comparison between performing the segmentation on a single projection or instead using the weighted sum over the three of them. In particular, values on the first column refer to the mean Dice obtained segmenting along each projection, while the second column to the Dice achieved with our approach.

Additionally, results reported in Table \ref{table_algo_comparison} show the final performance of our learning algorithm when trained on the whole train set and employing the entire data augmentation pipeline. Moreover a comparison with the most widely used classical methods is also carried out, in terms of segmentation quality and processing time. To test the neural network we employed a GPU NVIDIA GTX 970, while we used an Intel Xeon E5-2620 v4 CPU, working frequency 2.10GHz, with 16 cores and 32 threads, for all the other approaches.

\begin{table*}[]
\centering

       \begin{tabular}{llll}  
           \cmidrule{1-4}
             Label  		    				& Dice (\%) 	  			& FNR (\%)					& FPR (\%)		\\
           \midrule
      \rowcolor{lightgray} 0   		& $93.0 \pm 3.3$	 	& $0.5 \pm 0.3$	      	& $1.7 \pm 0.5$	      	\\
           		1     						& $94.0 \pm 2.3$	     & $2.8 \pm 3.3$	      	& $1.1 \pm 0.2$	      		 	    \\
           		2     						& $96.1 \pm 0.4$	     & $0.7 \pm 0.5$	      	& $0.8 \pm 0.1$	      		         \\
           		3     						& $95.7 \pm 1.7$	     & $4.4 \pm 2.4$	      	& $0.5 \pm 0.1$	      			     \\
           		4     						& $94.4 \pm 1.2$	     & $1.7 \pm 1.2$	      	& $1.1 \pm 0.3$	      				  \\
      \rowcolor{lightgray}ALL   	& $96.4 \pm 0.4$  	     & $0.1 \pm 0.1$	    		&$0.8 \pm 0.1$	      					\\
           \bottomrule
         \end{tabular} 
\caption{Comparison of the performance obtained using several data augmentation processes on reduced dataset. Refer to \ref{table_data_aug_settings} for the label meaning.}
\label{table_data_aug_comparison}
\end{table*}

\begin{table*}[]
\centering

       \begin{tabular}{lcc}  
           \multicolumn{3}{c}{Dice (\%)}                   \\
           \cmidrule{1-3}
              Label  & Single Projection  & Projections weighted mean  \\
           \midrule
           		0    &$89.9$  			&$93.0$  				\\
         		1    &$88.0$  			&$94.0$  				\\
           		2    &$95.6$  			&$96.1$  				\\
           		3    &$91.7$  			&$95.7$  				\\
           		4    &$92.1$  			&$94.4$  				\\
           		ALL  &$95.8$  			&$96.4$  		\\
           \bottomrule
         \end{tabular} 
\caption{Comparison of results obtained by testing over a single projection or three projections.}
\label{table_single_vs_all_projs}
\end{table*}

\begin{table*}[]
\centering

       \begin{tabular}{lllll}  
           \cmidrule{1-5}
             Method   		& Processing time (s)		& Dice (\%)  					& FNR 	(\%)							& FPR	(\%)		\\
           \midrule
           		AFNI      		& $117.7 \pm 53.3$		& $91.8 \pm 1.0$ 				& $2.8 \pm 0.2$	 				& $14.8 \pm 1.8$				\\
           		ANTS     		& $806.9 \pm 87.9$		& $\mathbf{95.2 \pm 0.4}$&$\mathbf{1.2 \pm 0.8}$	& $\mathbf{8.1 \pm 1.3}$ 	\\
           		BSE     			& $\mathbf{3.5 \pm 0.7}$	& $92.3 \pm 17.2$			& $5.5 \pm 10.9$ 				& $8.6 \pm 17.4$ 				\\
           		FSL     			& $23.4 \pm 11.3$			& $65.5 \pm 28.0$			& $16.1 \pm 23.7$ 				& $40.3 \pm 32.3$ 				\\
           		ROBEX     		& $66.5 \pm 2.7$			& $94.7 \pm 1.2$			& $3.4 \pm 1.3$ 					& $7.0 \pm 2.2$ 					\\
\rowcolor{lightgray}Our   & $\mathbf{4.5\pm 0.0}$ & $\mathbf{96.5\pm0.4}$				& $\mathbf{0.2\pm0.2}$					& $\mathbf{0.8\pm0.1}$					\\
           \bottomrule
         \end{tabular} 
\caption{Comparison of the performance obtained by each method on test set.}
\label{table_algo_comparison}
\end{table*}

\begin{figure*}[t]
      \centering
          \includegraphics[width=0.80\textwidth]{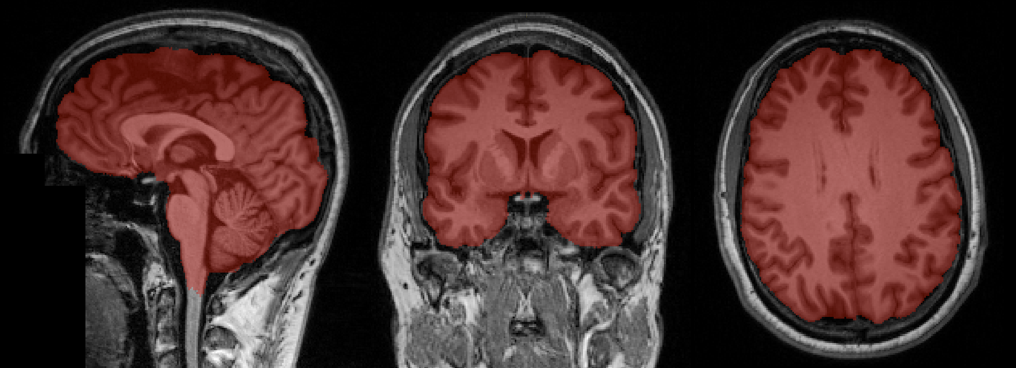}
      \caption{Example of brain extraction using the proposed CNN. Best viewed in color.}
      \label{fig_segmentation}
\end{figure*}

\section{Discussion}
As evidenced in Table \ref{table_data_aug_comparison}, the proposed neural network was able to obtain performances in line with the state of the art even if trained on a reduced data set. 

Furthermore, the results obtained from the study of the data augmentation pipeline show that it is possible to push the model to generalize much better even when we train on a limited number of cases. In fact, Table \ref{table_data_aug_comparison} highlights how all the analyzed operations contribute to improve the quality of the segmentation with respect to the baseline (Label 0). The additional weighted sum over the three projections makes the process even more robust (Table \ref{table_single_vs_all_projs}). 

As shown in Table \ref{table_algo_comparison}, no substantial improvement was introduced training on the whole dataset with respect to the case of reduced dataset (Dice of 96.5\% vs. 96.4\%). This means that no relevant variability was introduced in the additional data and consequently we believe that further improvement could be achieved only introducing new data from outside the NFBS dataset.

Table \ref{table_algo_comparison} puts in evidence that the final model of the network was able to obtain cutting-edge performance both in terms of computation time and quality of the segmentation. Additionally, since this tool does not require initial parameters setting, it is also easy to use and does not require any initial waste of time to find the optimal working configuration. Also, the time needed to skull-strip a brain scan is further reduced because $-$ thanks to the introduction of the synthetic bias in the data augmentation pipeline $-$ it is robust to changes in brain tissue intensity and no longer requires the preliminary correction of the magnetic field bias in the image.

In Figure \ref{fig_segmentation} an example of the obtained segmentation of the brain tissue is illustrated.

To conclude, despite the way in which we constructed the test set should help the analysis of the generalization capacity of the model, the used volumes still represent a limited number of cases of the global population. For this reason, although we experimented that the segmentation quality on different data seems to agree with the previous observations, a further quantitative analysis should be carried on. 

\section{Conclusion}
In this work we proposed and evaluated a new brain extraction method based on Convolutional Neural Networks. Moreover, we analyzed how the performance of the algorithm can be increased by a well designed data-augmentation pipeline. 

Although this study is still preliminar and certainly requires further investigation on different datasets, our model proved to have the potentialities to become a valid alternative to classical approaches since it demonstrated to be faster and more accurate. 

In conclusion, we underline that deep learning can play a relevant role in the skull-stripping process and can provide valid tools to be employed during most neuroimaging studies.

\bibliographystyle{unsrt} 
\bibliography{paper.bib}

\end{document}